\documentclass[10pt,twocolumn,letterpaper]{article}

\usepackage{iccv}
\usepackage{times}
\usepackage{epsfig}
\usepackage{graphicx}
\usepackage{amsmath}
\usepackage{amssymb}
\usepackage{color}

\DeclareMathOperator*{\softmax}{softmax}
\usepackage{multirow}
\usepackage{algorithmic}
\usepackage{bm}
\usepackage{bbm}
\usepackage{booktabs, multicol, multirow}
\def\signed #1{{\leavevmode\unskip\nobreak\hfil\penalty50\hskip2em
  \hbox{}\nobreak\hfil(#1)
  \parfillskip=0pt \finalhyphendemerits=0 \endgraf}}
\newsavebox\mybox

\usepackage[footnotesize]{subfigure}

\usepackage[lined,linesnumbered,ruled]{algorithm2e}

\SetCommentSty{mycommfont}
\DeclareMathOperator{\HACAN}{HACAN}

\usepackage[pagebackref=true,breaklinks=true,letterpaper=true,colorlinks,bookmarks=false]{hyperref}

\iccvfinalcopy 

\def\iccvPaperID{5927} 

\begin{document}

\title{Making History Matter: History-Advantage Sequence Training for Visual Dialog}

\author{Tianhao Yang$^{1}$ ~~~~~~~~Zheng-Jun Zha$^{1}$   ~~~~~~~~Hanwang Zhang$^2$\\
$^1$University of Science and Technology of China,
	$^2$Nanyang Technological University \\
	hanwangzhang@ntu.edu.sg;
}
\maketitle

\begin{abstract}
   We study the multi-round response generation in visual dialog, where a response is generated according to a visually grounded conversational history. Given a triplet: an image, Q\&A history, and current question, all the prevailing methods follow a codec (i.e., encoder-decoder) fashion in a supervised learning paradigm: a multimodal encoder encodes the triplet into a feature vector, which is then fed into the decoder for the current answer generation, supervised by the ground-truth. However, this conventional supervised learning does NOT take into account the \textbf{impact of imperfect history}, violating the conversational nature of visual dialog and thus making the codec more inclined to learn history bias but not contextual reasoning. To this end, inspired by the actor-critic policy gradient in reinforcement learning, we propose a novel training paradigm called \textbf{History-Advantage Sequence Training} (HAST). Specifically, we intentionally impose wrong answers in the history, obtaining an adverse critic, and see how the historic error impacts the codec's future behavior by \textbf{History Advantage} --- a quantity obtained by subtracting the adverse critic from the gold reward of ground-truth history. Moreover, to make the codec more sensitive to the history, we propose a novel attention network called \textbf{History-Aware Co-Attention Network} (HACAN) which can be effectively trained by using HAST. Experimental results on three benchmarks: VisDial v0.9\&v1.0 and GuessWhat?!, show that the proposed HAST strategy consistently outperforms the state-of-the-art supervised counterparts.
\end{abstract}
\begin{figure}[t]
\begin{center}
\includegraphics[width=1.0\linewidth]{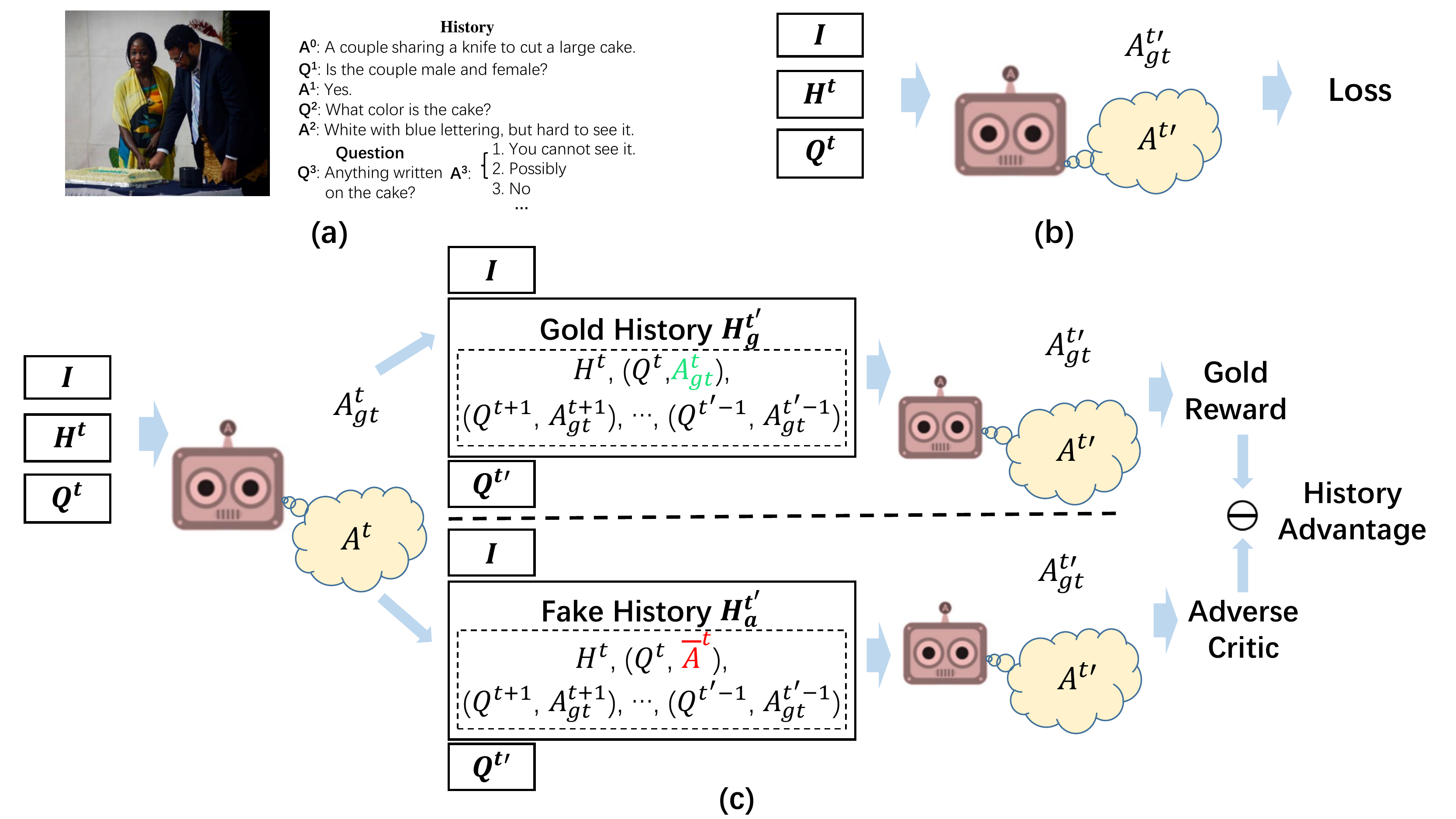}
\end{center}
    \caption{(a) A typical visual dialog task. Particularly, the initial answer $A^0$ denotes the given image captioning. (b) The conventional training process at round $t$: given an image $I$, a history $H^{t}$, and question $Q^t$, the loss is supervised by the ground-truth answer {\color{green}$A^t_{gt}$}. (c) The proposed History-Advantage Sequence Training (HAST) paradigm: the reward (\emph{i.e.} $-$ loss) is a History-Advantage, which is more focused on the impact caused by a wrong answer {\color{red}$\bar{A}^t$} to the future round $t'$, by comparing the difference between the Gold Reward from gold history $H^{t'}_g$ and the Adverse Critic from fake history $H^{t'}_a$.
    }
  \label{fig:1}
\end{figure}

\section{Introduction}
Visual dialog is one of the most comprehensive task for benchmarking the AI's comprehension of natural language grounded by a visual scene~\cite{das2017visual}. A good visual dialog agent should accomplish a complex series of reasoning sub-tasks: contextual visual perception~\cite{wu2017you,lu2017best,das2017human,seo2017visual,kottur2018visual,yang2016stacked,shih2016look}, language modeling~\cite{hochreiter1997long,chung2014empirical}, and co-reference resolution in dialog history~\cite{seo2017visual,kottur2018visual} (\eg, identify what is ``that''). Thanks to the end-to-end deep neural networks in their respective sub-tasks, state-of-the-art visual dialog systems can be built by assembling them into a codec framework~\cite{wu2017you,das2017visual,lu2017best}. The encoder encodes a triplet input --- history question-answer sentences, an image, and a current question sentence --- into a vector representation; then, the decoder fuses those vectors and decodes them into answer sentences (\eg, by generative language models~\cite{das2017visual,lu2017best,wu2017you} or discriminative candidate ranking~\cite{jain2018two,kottur2018visual}).

So far, one may identify that the key difference between Visual Dialog (VisDial) and the well-known Visual Question\&Answering (VQA)~\cite{antol2015vqa} is the exploitation of the \emph{history}. As shown in Figure~\ref{fig:1}(a) and Figure~\ref{fig:1}(b), VisDial can be cast into a multi-round VQA given additional language context of history question-answering pairs. Essentially, at each round, the response generated by the agent is ``thrown away'' and the history is artificially ``corrected'' by the ground-truth answers~\cite{das2017visual} for the next round. Note that this ground-truth history setting is reasonable because it steers the conversation to be evaluable; otherwise, any other response may digress the conversation into a never-ending open-domain chitchat~\cite{quarteroni2009designing,higashinaka2014towards}. However, we argue that \emph{by only exploiting the ground-truth history is ineffective for the codec training}. For example, the ground-truth answer only tells the model that ``\emph{you can not see it}'' is good, but neglect to show how badly other answers will impact the dialog. Therefore, the resultant model is easily over-grounded in the insufficient ground-truth data, but not learning visual reasoning~\cite{lake2017building}.

In this paper, we propose a novel training strategy that utilizes the history response in a more efficient way, that is, to make the codec model more sensitive to the history dialog such as co-reference resolution and context. As illustrated in Figure~\ref{fig:1}(c), we intentionally impose wrong answers in the ``tamperred'' history (\emph{e.g.} replacing ``\emph{White with blue lettering, but hard to see it.}'' with ``\emph{ Yellow}'' in the history) and see how the model behaves as compared to the ``gold'' history. Specifically, suppose that we are going to train a model at the $t$-th round, to gain more insights about the wrong answers, we maintain two parallel lines of dialogs: one with the ground-truth answer $A^t_{gt}$, and the other one with a probable mistake $\overline{A}_t$. Then, we run the two lines to a future round $t'$ (both of them are filled with ground-truth answers from $t+1$ to $t'$). Thus, we can collect two rewards at round $t'$: 1) \emph{Gold Reward} (GR): a conventional ground-truth history reward for answer $A^t_{gt}$, the larger the better , and 2) \emph{Adverse Critic} (AC): a proposed critic for fake history reward of $\overline{A}_t$, small AC implies large future impact by the mistake answer. Interestingly, their difference $\textrm{HA} = \textrm{GR} - \textrm{AC}$, which we call \textit{History-Advantage} (HA), will tell the model how to reward $A^t_{gt}$: if $\textrm{HA} > 0$, large $\textrm{HA}$ implies large impact of fake history, \ie, very small $\textrm{AC}$; thus, we need to strengthen the training signal of gold history with $A^t_{gt}$; if $\textrm{HA} \leq 0$, it implies the ineffectiveness of the gold history, \ie, the current model cannot accurately rank correct answer candidates; thus, the gradient for $A^t_{gt}$ will be in the opposite direction. In this way, we can collect a sequence of history-advantage training losses from $t+1$ to $T$. Therefore, we call the proposed training paradigm: History-Advantage Sequence Training (HAST).

Although the application scenario of HAST is regardless of specific codec models, for more effective training, we develop a novel codec dubbed: History-Aware Co-Attention (HACAN) encoder to address the essential co-reference and visual context in history encoding. In a nutshell, HACAN is a sequential model that contains two novel co-attention modules and one history-aware gate. Equipped with the the proposed History-Advantage Sequence Training (HAST), we achieve a new state-of-the-art single-model on the real-world VisDial benchmarks: 0.6792 MRR on VisDial v0.9, 0.6422 MRR on  VisDial v1.0, and 66.8\% accuracy on GuessWhat?!~\cite{de2017guesswhat}. We also achieve a top performing 0.5717 NDCG score on the official VisDial online challenge server. More ablative studies, qualitative examples, and detailed results are discussed in Section~\ref{sec:5}.

\begin{figure*}[t]
\begin{center}
\includegraphics[width=0.8\linewidth]{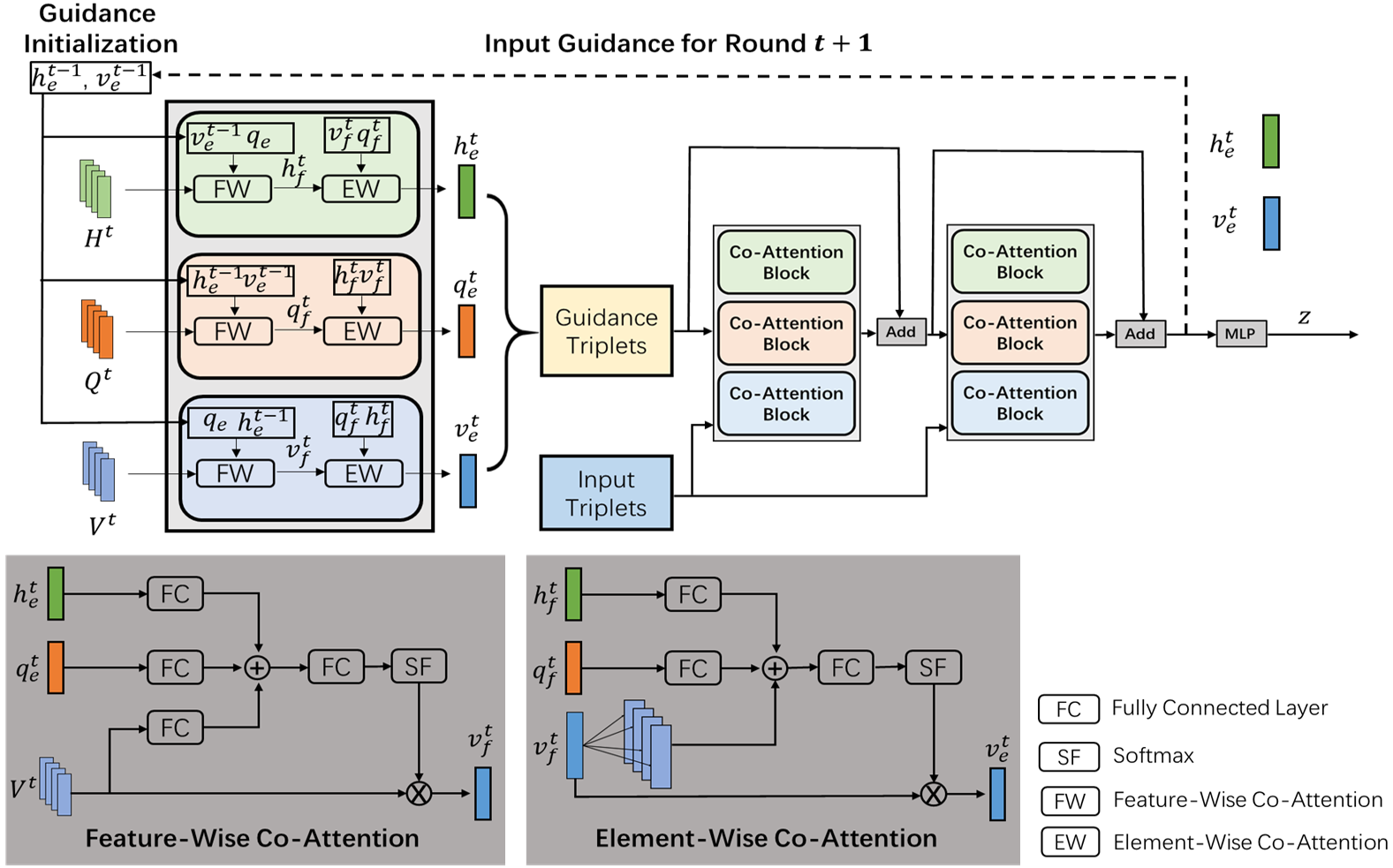}
\end{center}
    \caption{The framework of our proposed codec model. $\bm{H}^t$, $\bm{Q}^t$ and $\bm{V}^t$ are the input triplets (Section~\ref{sec:3.1}) extracted by CNNs and LSTMs. $t$ denotes the current round of the dialog. Feature-Wise Co-Attention and Element-Wise Co-Attention are applied as blocks in parallel to encode the input triplets and generate the guidances of the follow-up attention layer. The history-awareness, $\bm{h}_e^{t-1}$ and $\bm{v}_e^{t-1}$, are used to initialize the guidances in the first block (Section~\ref{sec:3.2}). The final outputs of the encoder update the history-awareness and feed to the decoder as well. The decoder finally generates the response and ranks candidate answer options (Section~\ref{sec:3.3}).
    }
\label{fig:model}
\end{figure*}

\section{Related Work}\label{sec 2}

\noindent\textbf{Visual Dialog.} Visual dialog is recently proposed in~\cite{das2017visual} and~\cite{de2017guesswhat}, which is a more challenging vision-language problem. Most vision-and-language problems are based on a single-round language interaction (\emph{e.g.} image caption~\cite{donahue2015long,fang2015captions,karpathy2015deep,anderson2017bottom} and visual question answering~\cite{anderson2017bottom,agrawal2016sort,antol2015vqa,das2017human,lu2016hierarchical}). On the contrary, visual dialog task involves a multi-round dialog which is more complex. Das \textit{et al.}~\cite{das2017visual} proposed a large-scale free-form visual dialog dataset, which consists of sequential open-ended questions and answers about arbitrary objects in the image. Another visual dialog task GuessWhat?! proposed by ~\cite{de2017guesswhat} focuses on a different aspect, which aims at object discovery with a set of yes/no questions. We apply the first setting in this paper. The proposed approaches for Visual Dialog are based on encoder-decoder structure, and can be categorized into three groups based on the design of encoder: (1) Fusion-based Models (LF~\cite{das2017visual}, HRE~\cite{das2017visual}, Sync~\cite{guo2019image}), the methods fuse image, question, and history features at different stages. (2) Attention-based Models (MN~\cite{das2017visual}, HCIAE~\cite{lu2017best}, CoAtt~\cite{lu2016hierarchical}), the methods
establish attention mechanisms over image, question and history. (3) Approaches that deal with visual reference resolution (AMEM~\cite{seo2017visual}, CorefNMN~\cite{kottur2018visual}, RvA~\cite{kang2019dual}, DAN~\cite{kang2019dual}), the methods focus on explicit visual co-reference resolution in visual dialog.

\noindent\textbf{Attention Mechanism.} Attention mechanisms are widely used in vision-and-language problems, and have achieved inspiring. For visual question answering, the attention-based model may attend to both the relevant regions in the image and the phrases in the questions. A number of approaches\cite{yang2016stacked, zhu2017structured, das2017human, fukui2016multimodal, shih2016look} have been proposed that apply question-based attention module on image features. Attention models~\cite{lu2016hierarchical, nam2016dual} which attend to the phrases or words in the questions are developed in later studies. We apply co-attention mechanism in our proposed model HACAN.

\noindent\textbf{Reinforcement Learning with Baseline.} Reinforcement learning with baseline is widely applied in language generation, \eg, image caption~\cite{rennie2017self} and visual dialog~\cite{lu2017best, wu2017you}. However, the historical response rounds in dialog are not considered as a sequence, which is in contrast our work focuses on. In particular, we follow the spirit of sequence training and design a history-advantage as a baseline to reward or penalize the future dialog rounds based on the current gold and fake history.


\section{Our Codec Model}\label{sec3}
In this section, we formally introduce the visual dialog task and describe the details of our proposed model. We follow the definition introduced by Das~\etal~\cite{das2017visual}. Given input as: 1) an image $I$, 2) a dialog history with a caption $A^0$ of the image and $t\!-\!1$ rounds of the dialog  $\{A^0,(Q^{1},A^{1}),...,(Q^{t-1},A^{t-1})\}$, where $(Q^{i},A^{i})$ is the $i$-th round of the ``ground-truth'' question and answer pair, 3) a follow-up question $Q^{t}$, and 4) a list of 100 candidate answer options $\{{A^{t}_{1}},...,{A^{t}_{100}}\}$ which contains one correct answer $A^{t}_{gt}$. The visual dialog model needs to sort the answer options and choose the right one when given the inputs. To perform response generation given the above task, as illustrated in Figure~\ref{fig:model}, our codec model includes three modules: 1) feature representation (Section~\ref{sec:3.1}), 2) the proposed History-Aware Co-Attention Network (HACAN) for the encoder (Section~\ref{sec:3.2}), and 3) a discriminative decoder for response generation by ranking (Section~\ref{sec:3.3}). 
\subsection{Feature Representation}\label{sec:3.1}

\noindent\textbf{Image Feature.} We follow bottom-up attention mechanism to extract region-based image features as in~\cite{anderson2017bottom}. 
We train Faster-RCNN~\cite{ren2015faster} based on the ResNet-101 backbone~\cite{he2016deep} on Visual Genome~\cite{krishna2017visual} dataset. We choose top-K regions from each image and encode the regions as the visual feature $\bm{V}^t$, where $t$ is the current round in the dialog. For fair comparison with some methods without region proposal network~\cite{ren2015faster}, we replace bottom-up attention mechanism with VGG model in the ablation study. More details are discussed in Section~\ref{sec:5}.

\noindent\textbf{Language Feature.} 1) Question Feature: We first embed the words of the follow-up question.  LSTM is applied to generate a sequence of hidden states. Words play different roles in the question. The operative words can tell the model the kind of the question and which instance attribute to consider. Especially in visual dialog task, history and the current question may have relationships and make word contribution mean more. As illustrated in Figure~\ref{fig:1}(a), the word ``\emph{written}'' in the current question has relationship with ``\emph{blue lettering}'' and helps the model focus on the last round in the history. Meanwhile, the last round reminds the model to pay more attention to the word ``\emph{written}''. From this, we use the whole hidden state sequence generated by LSTM instead of last hidden state as some prior works do, which is denoted as $\bm{Q}^t$. 2) Answer Feature: We apply another LSTM to the word embeddings of the candidates, and use the the whole hidden state sequence as the answer feature. 3) History Feature: Each round of the question and the answer in the history is concatenated into a long ``sentence''. Another LSTM is applied to each round of the history after word embedding. We use the last hidden state of each round as the question-answer pair feature. The history feature is denoted as $\bm{H}^{t}$.

\subsection{History-Aware Co-Attention Network}\label{sec:3.2}
We propose a novel attention-based model called History-Aware Co-Attention Network (HACAN) to encode the input features described above with the co-attention mechanism~\cite{wu2017you,lu2017best,lu2016hierarchical}. 

As illustrated in Figure~\ref{fig:model}, HACAN is composed by a sequence of attention blocks,  each of them contains two attention-based modules: Feature-Wise Co-Attention module (FCA) and Element-Wise Co-Attention module (ECA). Given the input triplet $\{\bm{V}^t, \bm{Q}^t, \bm{H}^{t}\}$, FCA aims to attend to the relevant features in one set of inputs with the guidances from the other two inputs. ECA takes the outputs of FCA as the inputs. It aims to activate the relevant elements and restrain the irrelevant ones with the guidances from the other two inputs before the final fusion. We now describe the two modules in details. 

\noindent\textbf{Feature-Wise Co-Attention (FCA).} We use additive attention function to compute the feature-wise attention, returning the attended feature as output. Without loss of generality, we take the FCA for $V$ as an example, which is denoted as $Attend_f(\bm{V}^t, \bm{g}^Q, \bm{g}^H)$. Take the visual feature of image regions $\bm{V}^t=\{\bm{v}_1,...\bm{v}_K\}$ as input, the visual attention is formulated as: 

\begin{equation}
    \bm{v}'_{i}=\tanh(W_{1}^{f}\bm{v}^t_{i}+W_{2}^{f}\bm{g}^Q+W_{3}^{f}\bm{g}^H),\label{equ:tanh}
\end{equation}
\begin{equation}
    \alpha_{i}=\softmax(W^{T}\bm{v}'_{i}),\label{equ:softmax}
\end{equation}
\begin{equation}
    \bm{v}^t_f=\sum_{i=1}^{K} \alpha_{i} \bm{v}_{i} \quad i=1,...,K,\label{equ:sum}
\end{equation}
where $\bm{g}^Q, \bm{g}^H \in \mathbb{R}^{d}$ are the guidances from $\bm{Q}^t$ and $\bm{H}^t$. $W_1^{f}, W_2^{f}, W_3^{f} \in \mathbb{R}^{d \times d}$, $W \in \mathbb{R}^{d \times 1}$ and $d$ is the feature dimension. As the formats of $\bm{Q}^t$ and $\bm{H}^t$ are consistent with $\bm{V}^t$, we can simply apply Eq.~\eqref{equ:tanh}-\eqref{equ:sum} to $Attend_f(\bm{Q}^t, \bm{g}^H, \bm{g}^V)$ and $Attend_f(\bm{H}^{t}, \bm{g}^V, \bm{g}^Q)$, and compute three modules in parallel.

\noindent\textbf{Element-Wise Co-Attention (ECA).} The outputs of FCAs are the sum of attended features. However, the features itself are not attended yet. We introduce an element-wise attention mechanism to attend the sum of features. It is worth noting that, each element in the feature (\emph{e.g.}, $\bm{v}_{i} \in \mathbb{R}^d$ in $\bm{V}^t$) is a response activation of the neural networks, and reflects some attributes of the instance in a sense. We apply the attention mechanism in an element-wise manner with the guidance from the other two inputs. It can be regarded as selecting the relevant semantic attributes and discarding the irrelevant ones based on the guidances from other domains. We take $Attend_e(\bm{v}^t_f, \bm{q}^t_f, \bm{h}^t_f)$ as an example, and ECA is formulated as:

 \begin{equation}
     \bm{V}^{e}=\tanh(W_{1}^{e}\otimes \\ 
     \bm{v}^t_f+(W_{2}^{e}\bm{q}^t_f)\bm{E}^T+(W_{3}^{e}\bm{h}^t_f)\bm{E}^T),\label{equ:tanh2}
 \end{equation}
 \begin{equation}
     \bm{v}^t_e=\sigma(W'^T \bm{V}^{e}) \odot \bm{v}^t_f,\label{equ:sig}
 \end{equation}
 where $W_{1}^{e} \in \mathbb{R}^{m}$ which is learnable and all elements are initialized with value 1, $W_{2}^e, W_{3}^e \in \mathbb{R}^{d \times d}$ and $W'\in \mathbb{R}^{m}$. $\bm{E} \in \mathbb{R}^{m}$ is a vector with all elements set to 1. $\otimes$ represents the outer product of vectors and $\odot$ represents the element-wise product. We use $W_{1}^{e}$ to broadcast $\bm{v}^t$ to \emph{m} times, and perform the attention function in parallel. It can be viewed as the additional-attention of multi-head attention in ~\cite{vaswani2017attention}.
 We compute $Attend_e(\bm{q}^t_f, \bm{h}^t_f, \bm{v}^t_f)$ and $Attend_e(\bm{h}^t_f, \bm{v}^t_f, \bm{q}^t_f)$ in the same way.
 
\noindent\textbf{Gated History-Awareness}. We observe that an ambiguous question often has relationship with its latest round in the history. From this, the encoding feature of the latest history $v^{t-1}_{e}, h^{t-1}_{e}$ is a good choice to initialize the guidances in FCAs, which can be viewed as history-awareness. However, sometimes the current question has nothing to do with its latest history. A simple solution is to apply a gate function to control the history-awareness, which is formulated as:

\begin{equation}
    \bm{q}_{s}^t = Attend_f(\bm{Q}^t, \bm{0}, \bm{0}), \label{equ:self attention}\\
\end{equation}
\begin{equation}
    \bm{o} = MLP([f^q_{g}(\bm{q}_{s}^t), f^h_{g}(\bm{h}_e^{t-1})]),
\end{equation}
\begin{equation}
    \begin{split}
        \bm{h}^{t-1}_e = \sigma(\bm{o})\bm{h}^{t-1}_e,  \quad \bm{v}^{t-1}_e = \sigma(\bm{o})\bm{v}^{t-1}_e, \\
    \end{split}
\end{equation}
 where $\bm{q}^t_s$ is self-attention of the current question, $[\cdot]$ is a concatenation operation and $\sigma(\bm{o})$ is the gate value. With the gated history-awareness, each first FCA is guided by its latest history feature and update the attended features efficiently. For the follow-up modules, we take the outputs of ECAs as the guidance inputs in Eq.~\eqref{equ:tanh}, so we can connect two attention modules recurrently and update the encoding feature $\bm{v}^t_e, \bm{h}^t_e, \bm{q}^t_e$, as illustrated in Figure ~\ref{fig:model}. The shortcut connections in the layers help the model consider different levels in multi-hop visual reasoning. The outputs of final round of attention blocks, $\bm{v}^t_e, \bm{h}^t_e, \bm{q}^t_e$, are used to generate the answer response.

\subsection{Response Generation}\label{sec:3.3}
Now we introduce how to generate the answers for the visual dialog task. We concatenate the three features $\bm{v}^t_e, \bm{h}^t_e, \bm{q}^t_e$ together and use a linear transform followed by a tangent activation: 
\begin{equation}
\bm{z}=\tanh(W_{e}[\bm{v}^t_e, \bm{h}^t_e, \bm{q}^t_e]), \label{equ:fusion}
\end{equation}
where $[,]$ is a concatenation operation.
We encode candidate answer features using a self-attention mechanism. The self-attention mechanism is formulated like Eq.~\eqref{equ:self attention}. We dot product the candidate answer features and $\bm{z}$ to calculate the similarities. We sort the answer candidates by the similarities and choose the top one with highest similarity as the prediction.

In the task of GuessWhat?!, the information of answer candidates are the localizations and categories of the objects. We concatenate the localizations and categories and embed them with a linear transform to obtain the answer option features. We also calculate the similarities of answer features and the final encoding features by dot product.
\section{History-Advantage Sequence Training}\label{sec:4}
HACAN described in Section~\ref{sec:3.2} encodes the ``ground truth'' triplet and generates the responds. However, by only using the conventional supervised learning, HACAN does not take into account the contribution of ``gold'' history and the impact of imperfect history, which makes the model history-aware and explore semantic information in the history. To disentangle the individual contribution of a specific round in the history and make the codec model more sensitive to the history, an intuitive solution is to replace the default answer of the specific round in the history with imperfect answers. To this end, besides utilizing ``gold'' history, we intentionally impose wrong answers chosen by the model to generate the ``tamperred'' history, and see how much better the model performs with ``gold'' history than the one with ``tamperred'' history. We first describe the policy gradient for visual dialog in Section~\ref{sec:4.1}, and then we describe the history-advantage in the policy gradient in Section~\ref{sec:4.2}. Finally, we briefly introduce the training process in Section~\ref{sec:4.3}.

\subsection{Policy Gradient for Visual Dialog}\label{sec:4.1}
We transform visual dialog task to a simple two-step decision-making game: in step 1, given an image \emph{I}, $t$-round history with caption $\{A^0,(Q^{1},A^{1}),...,(Q^{t-1},A^{t-1})\}$ and the follow-up question $Q^t$, the agent needs to choose an answer from the candidates. In step 2, besides the inputs above and the choice of the agent, a judge is given the remaining rounds from $t+1$ to $t'$, and needs to mark by answering the current question $Q^{t'}$ (t' is anywhere from $t+1$ to 10) correctly. The goal of the game is to choose the right answer of $Q^{t'}$, which is the same as visual dialog task described in Section~\ref{sec3}. The difference is the choice of $Q^{t}$ may be imperfect and impact the score of the game. We formulate the game as a decision-making problem. The action space is the answer candidates denoted as $\{{A^{t}_{1}},...,{A^{t}_{100}}\}$ and the state space is $\{\bm{V}^{t'},\bm{Q}^{t'},\bm{H}^{t'}\}$. 


 
\subsection{History-Advantage}\label{sec:4.2}
 We use the metric of the visual dialog task to compute the reward (\emph{e.g.} MRR). We denote the gold reward (GR) as $R(\bm{V}^{t'},\bm{Q}^{t'},\bm{H}^{t'}_{g})$ and use $R(\bm{V}^{t'}, \bm{Q}^{t'}, \bm{H}^{t'}_{a,i})$ to represent the adverse critic (AC) when the agent choose the $i$-th negative answer in round $t$. As illustrated in Figure ~\ref{fig:1}(c), their difference $\textrm{GR} - \textrm{AC}$ can reflect the influence of the ``gold'' round $t$ when answering $Q^{t'}$. We definite history-advantage of the ``gold'' round $t$ as:
 
 \begin{equation}
    \begin{split}
     GR-AC = A(t') = A(\bm{V}^{t'},\bm{Q}^{t'},\bm{H}^{t'}) = \\
     R(\bm{V}^{t'},\bm{Q}^{t'},\bm{H}^{t'}_{g})
     - \sum p(\bm{\bar{A}}^{t}_{i})R(\bm{V}^{t'},\bm{Q}^{t'},\bm{H}^{t'}_{a,i}), \label{equ:advantage}
    \end{split}
 \end{equation}
where $A(t')$ can estimate the contribution of the ``gold'' answer of round $t$ on answering $Q^{t'}$. It plays a similar role as the ``\emph{advantage}'' in actor-critic methods, and AC is a baseline in the history-advantage. When the total adverse critic is lower than gold reward, $A(t')$ is positive. It means the mistake does impact negatively on the future and the influence of the ``gold'' round is positive. The baseline can reduce the variance of gradient estimation in the training part as well and help the model take the contribution of the history into account. As $t'$ is anywhere from $t+1$ to 10, we compute $A(t')$ with different value of $t'$, which can reduce the variance of gradient estimation as well. The sum of $A(t')$ can also be viewed as the contribution of the ``gold'' round in the whole dialog.

\subsection{Training}\label{sec:4.3}
Inspired by the policy gradient theorem~\cite{sutton2000policy}, the history-advantage gradient can be simply denoted as:
 \begin{equation}\label{equ:J3}
     \nabla_{\theta} J_{g}=\nabla_{\theta}\log \bm{p}(A^t_{gt};\theta) \cdot(\frac{1}{10-t}\sum_{t'=t+1}^{10}A(t')),
 \end{equation}

Following previous policy gradient works that use a supervised pre-training step as model initialization, we train our model with two-stage training. In the first training stage, we use a metric-learning multi-class N-pair loss $\mathcal{L}_{np}$~\cite{sohn2016improved,lu2017best}.

In HAST, we denote the final gradient as:

\begin{equation}
    \nabla_{\theta} J= \nabla_{\theta} J_{g} - \alpha \nabla_{\theta} \mathcal{L}_{np},
     \label{equ:gra}
\end{equation}
where we incorporate N-pair loss (weighted by a trade-off $\alpha$) for an end-to-end training.
The whole training process is reviewed in Algorithm.~\ref{alg1}.

\begin{algorithm}
\caption{Discriminative Model with History-Advantage Sequence Training} 
\label{alg1}
\begin{algorithmic}[1]
\REQUIRE Supervised Pre-trained Model $\HACAN$
\FOR{Round $t=1,\cdots,T-1$}
    \STATE Compute $\{ p(\bar{A}^t_{1}),\cdots,p(\bar{A}^t_{N-1})\} \sim \HACAN(\bm{V}^t, \bm{Q}^t, \bm{H}^t)$\\
    where N is the number of answer candidates.
    \FOR{$t'=t+1,\cdots,T$}
        \FOR{$i=1,\cdots,N-1$}
            \STATE Fake the history $H^{t'}$ with $\bar{A}^{t}_{i}$
            \STATE Compute $R(\bm{V}^{t'},\bm{Q}^{t'},\bm{H}^{t'}_{g})$ and $R(\bm{V}^{t'}, \bm{Q}^{t'}, \bm{H}^{t'}_{a,i})$ 
            using $\HACAN$
        \ENDFOR
    \ENDFOR
    \STATE Compute the gradient $\nabla_{\theta} J$ with Eq.~\eqref{equ:gra}
    \STATE $\theta \leftarrow \theta + \delta\nabla_{\theta}J$
\ENDFOR
\end{algorithmic}
\end{algorithm}

\section{Experiments}\label{sec:5}
In the following we evaluate our proposed approach on three visual dialog datasets, VisDial v0.9~\cite{das2017visual}, VisDial v1.0~\cite{das2017visual} and GuessWhat?!~\cite{de2017guesswhat}. We first present the details about the datasets, evaluation metrics and the implementation details. Then we provide qualitative results and compare our methods with the state-of-the-art models.
\subsection{Datasets}
VisDial v0.9~\cite{das2017visual} contains about 123k image-caption-dialog
tuples. The images are all from MS COCO~\cite{lin2014microsoft} with multiple objects. The dialog of each image has 10 question-answer pairs, which were collected by pairing two people on Amazon Mechanical Turk to chat with each other about the image. Specifically, the ``questioner'' is required to ``imagine the scene better'' by sequentially asking questions about the hidden image. The ``answerer'' then observes the picture and answers questions. 

VisDial v1.0~\cite{das2017visual} is an extension of VisDial v0.9~\cite{das2017visual}. Images for the training set are all from COCO train2014 and val2014. The dialogs in validation and test sets were collected on about 10k COCO-like images from Flickr. The test set is split into two parts, 4k images for test-std and 4k images for test-challenge. Answers are already provided for the train and val set, but not in the test set. For the test-std and test-challenge phases, the results must be submitted to the evaluation server. 

We also evaluated our proposed model on GuessWhat?! dataset~\cite{de2017guesswhat}. The dataset contains 67k images collected from MS COCO~\cite{lin2014microsoft} and 155k dialogs including about 820k question-answer pairs. The guesser game in GuessWhat?! is to predict the correct object in object options through a multi-round dialog.

\linespread{0.65}
\begin{table*}
\begin{center}
\small
\begin{tabular}{
cccccccccccc}
\toprule
\multirow{2}{*}{} & \multicolumn{5}{c}{VisDial v1.0(test-std)} & \multicolumn{5}{c}{VisDial v0.9(val)}  \\
\cmidrule(r){2-7} \cmidrule(r){8-12} 
&  NDCG      &  MRR   &   R@1   &   R@5   &R@10    &  Mean   
&  MRR   &   R@1   &   R@5   &R@10    &  Mean  \\
\midrule
LF w/o RPN~\cite{das2017visual}    &0.4531 &0.5542 &40.95 &72.45 &82.83 &5.95 &0.5807 &43.82 &74.68 &84.07 &5.78         \\
HRE~\cite{das2017visual}   &0.4546 &0.5416 &39.93 &70.45 &81.50 &6.41 &0.5846 &44.67 &74.50 &4.22  &5.72         \\
MN~\cite{das2017visual}    &0.4750 &0.5549 &40.98 &72.30 &83.30 &5.92 &0.5965 &45.55 &76.22 &85.37 &5.46         \\
HCIAE~\cite{lu2017best}   &- &- &- &- &- &-                        &0.6222 &48.48 &78.75 &87.59 &4.81         \\
AMEM~\cite{seo2017visual}  &- &- &- &- &- &-                          &0.6227 &48.53 &78.66 &87.43 &4.86         \\
CoAtt~\cite{lu2016hierarchical} &- &- &- &- &- &-                          &0.6398 &50.29 &80.71 &88.81 &4.47         \\
CorefNMN~\cite{kottur2018visual}  &0.5470 &0.6150 &47.55 &78.10 &88.80 &4.40 &0.6410 &50.92 &80.18 &88.81 &4.45     \\
RvA w/o RPN~\cite{niu2018recursive} &0.5176 &0.6060 &46.25 &77.88 &87.83 &4.65 &0.6436 &50.40 &81.36 &89.59 &4.22   \\
Ours w/o RPN   &0.5281 &0.6174 &47.91 &78.59 &87.81 &4.63 &0.6451 &50.72 &81.18 &89.23 &4.32\\
\midrule
LF~\cite{das2017visual}    &0.5163	&0.6041	&46.18	&77.80	&87.30	&4.75 &- &- &- &- &-                        \\
RvA~\cite{niu2018recursive}   &0.5559 &0.6303 &49.03 &80.40 &89.83 &4.18 &0.6634 &52.71 &82.97 &\textbf{90.73} &\textbf{3.93}         \\
Sync\cite{guo2019image} &0.5732  &0.6220  &47.90 &80.43 &\textbf{89.95} &\textbf{4.17}    &- &- &- &- &-  \\
DAN~\cite{kang2019dual}   &\textbf{0.5759} &0.6320 &49.63 &79.75 &89.35 &4.30 &0.6638 &53.33 &82.42 &90.38 &4.04         \\
\midrule  
Ours &0.5717  &\textbf{0.6422}  &\textbf{50.88} &\textbf{80.63} &89.45 &4.20   &\textbf{0.6792} &\textbf{54.76} &\textbf{83.03} &90.68 &3.97                            \\
\bottomrule
\end{tabular}
\end{center}
\caption{Retrieval performance of discriminative models on the test-standard split of VisDial v1.0 and the validation set of VisDial v0.9. RPN indicates the usage of region proposal network.}
\label{table:v1.0test}
\end{table*}

\subsection{Evaluation Metrics}
For VisDial v0.9, we followed the evaluation protocol established in~\cite{das2017visual} and used the retrieval setting to evaluate the responses at each round in the dialog. Specifically, for each question we sorted the answer options and used Recall@$k$, mean reciprocal rank (MRR) and mean rank of the ground truth answer to evaluate the model. Recall@$k$ is the percentage of questions for which the correct answer option is ranked in the top $k$ predictions of a model. Mean rank is the average rank of the ground truth answer option. Mean reciprocal rank is the average of 1/rank of the ground truth answer option. For the test set of VisDial v1.0, we also evaluated our model using the newly introduced normalized discounted cumulative gain (NDCG). NDCG is invariant to the order of options with identical relevance and to the order of options outside of the top $K$, where $K$ is the number of answers marked as correct by at least one annotator. For GuessWhat?! dataset, we used classification accuracy to evaluate our model.

\subsection{Implementation Details}
\noindent\textbf{Language Processing.} We first tokenized the questions and answers using the Python NLTK toolkit and constructed a vocabulary of words that appear at least 5 times in the training split. All the words were embedded to a 300-dimension vector initialized by pre-trained GloVe~\cite{pennington2014glove} embeddings. The LSTMs of the question and history were two-layered, while they were one-layered for the answers. The hidden states in all LSTM were 512-d.

\noindent\textbf{Training Details.} We pretrained our codec using the supervised training for 15 epochs before starting HAST. We used Adam optimizer~\cite{kingma2014adam} and started the supervised training with the base learning rate of $1\!\times\!10^{-3}$ and decreasing to $5\!\times\!10^{-4}$ after 10 epochs. In HAST, the base learning rate was $1\!\times\!10^{-4}$, and decayed every 5 epochs with an exponential rate of 0.5. We set the hyper-parameter $m$ to 16, and $W_{1}^{e} \in \mathbb{R}^{16}$ in Eq.~\eqref{equ:tanh2}. We set $\alpha$ to 1 in Eq.~\eqref{equ:gra}.

In Eq.~\eqref{equ:advantage}, we needed to sum over all the negative answer candidates to calculate the advantage for HAST. It cost quite a lot of time to train (about 99 evaluations for each incorrect answers). However, we noticed that only a few negative answers have non-ignorable probabilities. For reducing the time cost, we made a trade-off between the accuracy and the speed and summed over the top-5 negative answers chosen by our model. The experiment showed that we saved 95\% training time with a slight performance drop.

\begin{figure}[t]
\begin{center}
\includegraphics[width=1.0\linewidth]{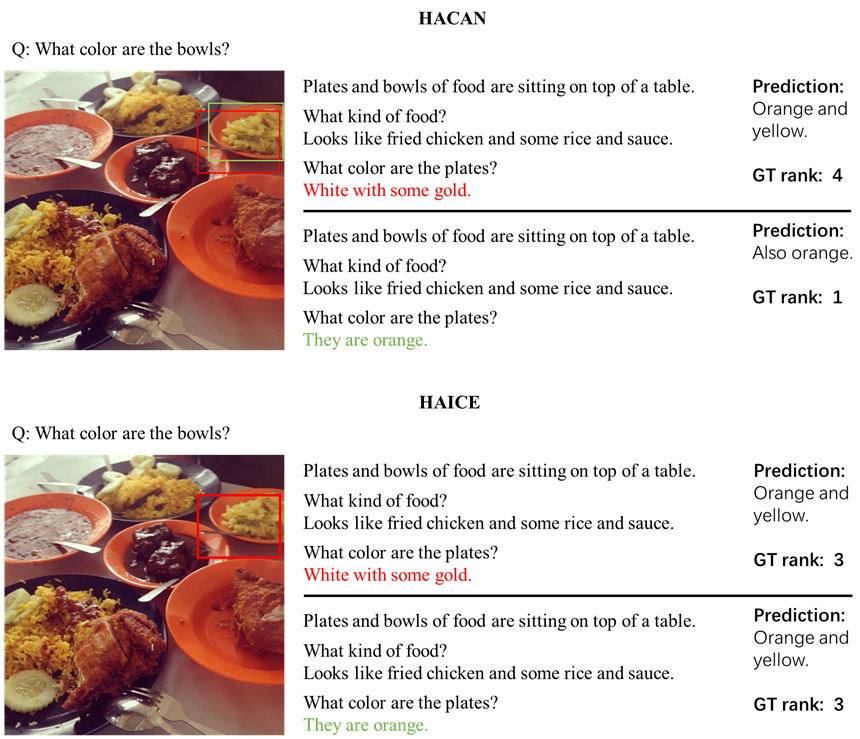}
\end{center}
   \caption{Qualitative results on VisDial dataset. We visualize the behaviors of our model and HCIAE with historical changes. Incorrect history and the region chosen by model with “tamperred” history are marked with red. GT rank denotes the rank of ground-truth answer in the sorted list.}
\label{fig:sample1}
\end{figure}
\begin{figure}[t]
\begin{center}
\includegraphics[width=1.0\linewidth]{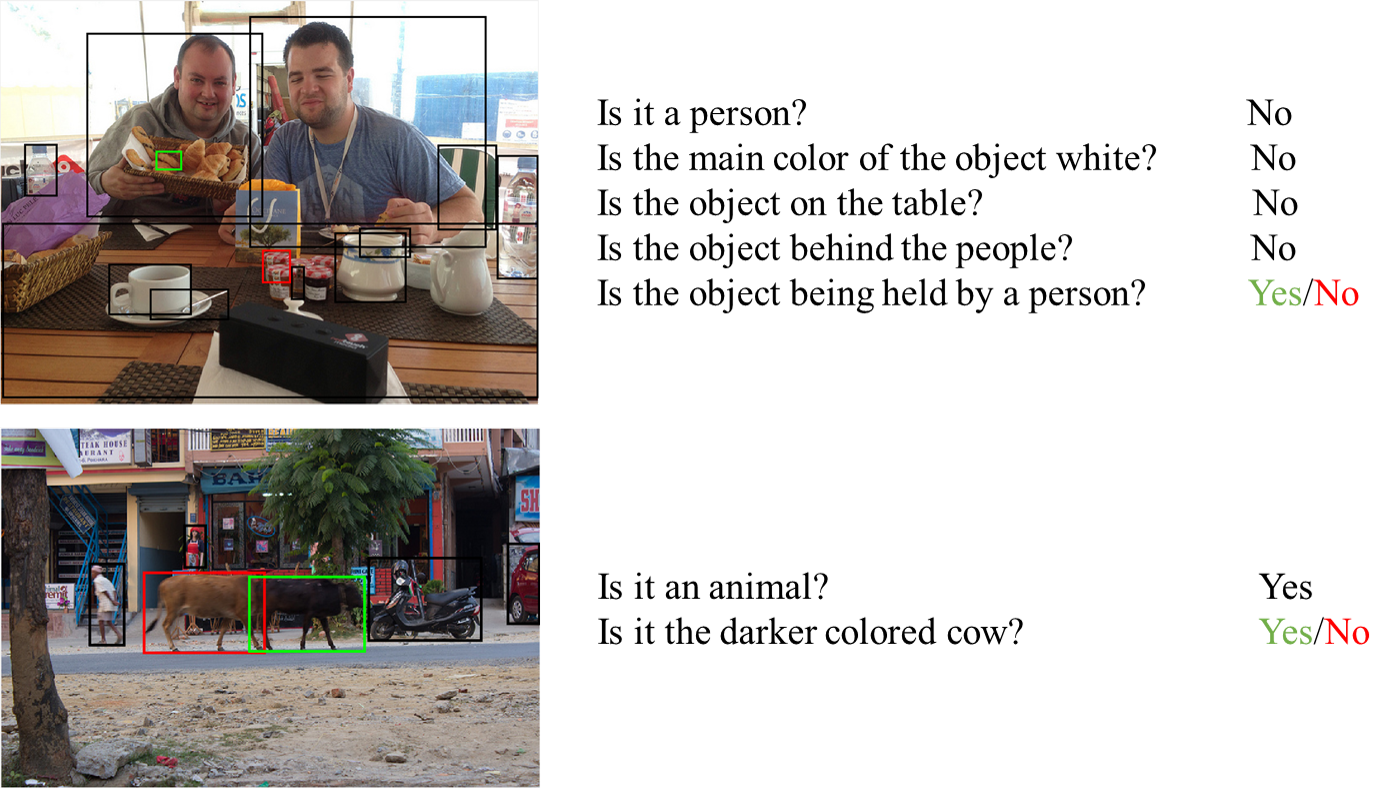}
\end{center}
   \caption{Qualitative results of our model on GuessWhat?!. The green bounding boxes highlight the right predictions with ’gold’ histories. The red bounding boxes highlight the wrong predictions with ’tamperred’ histories. }
\label{fig:sample2}
\end{figure}

\subsection{Ablative Study}
\noindent \textbf{Components in HACAN.} We present a few variants of our model to verify the contribution of each component:
\textbf{(1)}~HACAN w/o FCA is a baseline without FCA and ECA. Image features are guided by the question, the attention weight is computed by Eq.~\eqref{equ:tanh} with only question guidance.
~\textbf{(2)} HACAN w/o ECA-1 is a model with one FCA and Gated History-Awareness. The model has no ECA.
~\textbf{(3)} HACAN w/o ECA-16 is a model with one FCA, followed by one ECA with $m=1$ in Eq.~\eqref{equ:sig}. The ECA is not multi-head.
~\textbf{(4)} HACAN w/o RS is a model with one FCA, followed by one ECA with $m=16$. The numbers of multi-head is 16.
~\textbf{(5)} HACAN is a whole model that we stack two attention modules three times in a residual manner.

The first and second row in Table~\ref{table:ab_modules} show that, HACAN w/o FCA uses question-guided image feature (like the style in VQA task), but is not history-aware. Benefiting from FCA and Gated History-Awareness, the model takes history-aware features and achieves approximately 3.5\% improvements on MRR, more in line with the conversational nature of visual dialog. In HACAN w/o ECA-16, two types of attention modules guide the model not only which features to focus (\emph{e.g.} feature-level attended histories), but also the attended attributes of features, advancing the attention mechanism to a more fine-grained level. The hyper-parameter $m$ can be viewed as the number of heads in multi-head attention~\cite{vaswani2017attention}, and improves the performance by approximately 0.5 point on MRR. We stacked two attention modules up to three layers in a residual manner to obtain multi-level abstraction of history-aware and question-aware features, this achieved the best performance in our ablative experiments. We believe the ablative experiments demonstrate that: (1) different from the VQA task, the model for visual dialog task relies on history-aware representations. (2) FCA and ECA are efficient to compute both reliable question-aware and history-aware visual attention.

\noindent \textbf{Effectiveness of HAST.} We did ablative experiments to demonstrate the effectiveness of HAST. Results are showed in Table~\ref{table:ab_gcst} and Table~\ref{table:ab_gcst2}. Our model achieves approximately another 1 point improvement on R@1 and 0.5\% on MRR on VisDial dataset. Beyond that, we applied HAST to some additional ablative models with official codes\footnote{https://github.com/jiasenlu/visDial.pytorch}. We denote HCIAE-D-MLE~\cite{lu2017best} and HCIAE-D-NP-ATT~\cite{lu2017best} as HCIAE-M and HCIAE. The encoder models are attention-based. Interestingly, we found that both models achieve improvements with our proposed HAST. The results validate that: taking the impact of imperfect history into account, the history-aware models explore the contributions of ``gold'' history and deal with the relationship between visual and linguistic representations better.

The qualitative results shown in Figure~\ref{fig:sample1} and Figure~\ref{fig:sample2} demonstrate the following advantages of our HACAN model with HAST: 

\noindent \textbf{History Sensitive.} Our model is sensitive to the history. In Figure~\ref{fig:sample1}, the upper half part shows that, with historical changes, our model HACAN generates different responses. In more details, with the contribution of ``gold'' history, the ``ground truth'' answer achieves a higher rank order. Thanks to FCA and ECA, the visual attention influenced by historical changes is more precise when given the ``gold'' history. Rather, with historical changes, HCIAE behaves the same and does not benefit from the contribution of ``gold'' history.

\noindent \textbf{Reliable Contextual Reasoning.} Our model addresses contextual reasoning reliably using HAST. Focusing more on the influences of the different histories in the dialog, HACAN learns more contextual reasoning. In Figure~\ref{fig:sample2}, with different facts, HACAN chooses the corresponding regions in the image correctly. 

\subsection{Comparison with the State-of-the-art}
We compared our model with the state-of-the-art methods on VisDial v0.9 and v1.0. Early methods use grid-based CNN(\emph{e.g.} VGG-16~\cite{simonyan2014very}) features. For fair comparison, we replaced the bottom-up attention features~\cite{anderson2017bottom} with ImageNet pre-trained VGG-16 features. The upper half of Table~\ref{table:v1.0test} reports the results of the methods with the VGG-16 features, and the bottom half reports the results with bottom-up attention features. 

\noindent \textbf{Compared Methods.} The state-of-the-art methods can be categorized into three groups based on the design of encoder: (1) Fusion-based Models (LF~\cite{das2017visual}, HRE~\cite{das2017visual}, Sync~\cite{guo2019image}). (2) Attention-based Models (MN~\cite{das2017visual}, HCIAE~\cite{lu2017best}, CoAtt~\cite{lu2016hierarchical}). (3) Approaches that deal with visual reference resolution in VisDial (AMEM~\cite{seo2017visual}, CorefNMN~\cite{kottur2018visual}, RvA~\cite{kang2019dual}, DAN~\cite{kang2019dual})

Our method outperforms the state-of-the-art methods across most of the metrics. Specifically, our method achieves more than 1 point improvement on R@1, and 1\% increase on MRR. We also achieve a new state-of-the-art single-model on the official VisDial online challenge server\footnote{https://evalai.cloudcv.org/web/challenges/challenge-page/103/leaderboard/298}.
Furthermore, we conducted supplementary experiments on the guesser task of GuessWhat?!. Table~\ref{table:guesswhat} shows that our method is comparable to the state-of-the-art methods.

\linespread{0.65}
\begin{table}
\begin{center}
\scalebox{0.9}{
\begin{tabular}{lccc}
\toprule  
Model& Train err& Val err& Test err\\
\midrule  
LSTM~\cite{de2017guesswhat}& 27.9\% & 37.9\% &38.7\%\\
HRED~\cite{de2017guesswhat}& 32.6\% & 38.2\% &39.0\%\\
LSTM+VGG~\cite{de2017guesswhat}& \textbf{26.1}\% & 38.5\% &39.2\%\\
HRED+VGG~\cite{de2017guesswhat}& 27.4\% & 38.4\% &39.6\%\\
ATT~\cite{deng2018visual}& 26.7\% & 33.7\% &34.2\%\\
\midrule  
Ours& \textbf{26.1}\% & \textbf{32.3}\% &\textbf{33.2}\%\\
\bottomrule 
\end{tabular}}
\end{center}
\caption{Results on the guesser game of GuessWhat?!.}
\label{table:guesswhat}
\end{table}

\begin{table}
\begin{center}
\scalebox{0.8}{
\begin{tabular}{lccccc}
\toprule
Model  & MRR& R@1 &R@5 &R@10 &Mean \\
\midrule  
HACAN w/o FCA   & 0.5837& 44.52& 74.77& 84.84& 5.56\\
HACAN w/o ECA-1   &0.6181& 48.29& 78.23& 87.76& 4.77\\
HACAN w/o ECA-16  & 0.6285& 49.26& 79.41& 88.72& 4.53\\
HACAN w/o RS  & 0.6323& 49.61& 79.96& 89.05& 4.40\\ 
HACAN               &\textbf{0.6391} &\textbf{50.44} &\textbf{80.67} &\textbf{89.71} &\textbf{4.32}\\
\bottomrule 
\end{tabular}}
\end{center}
\caption{Performance of ablative models on the validation set of VisDial v1.0.}
\label{table:ab_modules}
\end{table}

\begin{table}
\begin{center}
\scalebox{0.75}{
\begin{tabular}{lccccc}
\toprule  
Model  & MRR& R@1 &R@5 &R@10 &Mean \\
\midrule  
HCIAE-M w/o HAST~\cite{lu2017best}& 0.6156 &47.67& 78.50& 87.54& 4.68\\
HCIAE-M~\cite{lu2017best}& 0.6177& 47.95& 78.70& 87.97& 4.61\\
\midrule  
HCIAE w/o HAST~\cite{lu2017best}& 0.6227& 48.58& 79.19& 88.16& 4.58\\
HCIAE~\cite{lu2017best}& 0.6243& 48.71& 79.14& 88.72& 4.53\\ 
\midrule  
HACAN w/o HAST   &0.6391 &50.44 &80.67 &89.71 &4.32 \\
HACAN &\textbf{0.6445} &\textbf{51.20} &\textbf{80.76} &\textbf{89.92} &\textbf{4.25}\\
\bottomrule 
\end{tabular}}
\end{center}
\caption{Performance of ablative models on the validation set of VisDial v1.0. HAST indicates the usage of History-Advantage Sequence Training.}
\label{table:ab_gcst}
\end{table}

\linespread{0.65}
\begin{table}
\begin{center}
\scalebox{0.8}{
\begin{tabular}{lccc}
\toprule  
Model& Train err& Val err& Test err\\
\midrule  
HRED w/o HAST~\cite{de2017guesswhat}& 32.6\% & 38.2\% &39.0\%\\
HRED~\cite{de2017guesswhat}& 31.8\% & 37.7\% &38.4\%\\
\midrule
HRED+VGG w/o HAST~\cite{de2017guesswhat}& 27.4\% & 38.4\% &39.6\%\\
HRED+VGG~\cite{de2017guesswhat}& 26.8\% & 37.7\% &38.9\%\\
\midrule  
HACAN w/o HAST& 26.9\% & 33.6\% &34.1\%\\
HACAN& \textbf{26.1}\% & \textbf{32.3}\% &\textbf{33.2}\%\\
\bottomrule 
\end{tabular}}
\end{center}
\caption{Performance of ablative models on the guesser game of GuessWhat?!. HAST indicates the usage of History-Advantage Sequence Training.}
\label{table:ab_gcst2}
\end{table}

\section{Conclusion}
In this paper we develop a codec model equipped with History-Aware Co-Attention Network (HACAN) for the visual dialog task. HACAN contains Feature-Wise Co-Attention module and Element-Wise Co-Attention module to address the co-reference and visual context in question and history encoding. We propose a novel training strategy dubbed History-Advantage Sequence Training (HAST) that utilizes the history response to make the codec model more sensitive to the history dialog. Extensive experiments on the real-world datasets, VisDial and GuessWhat?!, achieve a new state-of-the-art single-model on the benchmarks.


{\small
\bibliographystyle{ieee}
\bibliography{egbib}
}

\end{document}